\documentclass{article}
\usepackage{arxiv}
\usepackage[utf8]{inputenc}
\usepackage{graphicx}
\usepackage{hyperref}
\usepackage{amsmath}
\usepackage{amssymb}
\usepackage{subcaption}

\title{SimQ-NAS: Simultaneous Quantization Policy and Neural Architecture Search}
\author{Sharath Nittur Sridhar \thanks{Authors have equal contribution.} \\
    \small{Intel Labs, Intel Corporation} \\
    \small{sharath.nittur.sridhar@intel.com} \\\And
    Maciej Szankin $^{*}$ \\
    \small{Intel Labs, Intel Corporation} \\
    \small{maciej.szankin@intel.com} \\\And
    Fang Chen \\
    \small{University of California, Merced} \\
    \small{fchen20@ucmerced.edu} \\\And
    Sairam Sundaresan \\
    \small{Intel Labs, Intel Corporation} \\
    \small{sairam.sundaresan@intel.com} \\\And
    Anthony Sarah \\
    \small{Intel Labs, Intel Corporation} \\
    \small{anthony.sarah@intel.com} \\\
}

\begin{document}
\maketitle

\begin{abstract}
Recent one-shot Neural Architecture Search algorithms rely on training a hardware-agnostic super-network tailored to a specific task and then extracting efficient sub-networks for different hardware platforms. Popular approaches separate the training of super-networks from the search for sub-networks, often employing predictors to alleviate the computational overhead associated with search. Additionally, certain methods also incorporate the quantization policy within the search space.  However, while the quantization policy search for convolutional neural networks is well studied, the extension of these methods to transformers and especially foundation models remains under-explored. In this paper, we demonstrate that by using multi-objective search algorithms paired with lightly trained predictors, we can  efficiently search for both the sub-network architecture and the corresponding quantization policy and outperform their respective baselines across different performance objectives such as accuracy, model size, and latency. Specifically, we demonstrate that our approach performs well across both uni-modal (ViT and BERT) and multi-modal (BEiT-3) transformer-based architectures as well as convolutional architectures (ResNet). For certain networks, we demonstrate an improvement of up to $4.80x$ and $3.44x$ for latency and model size respectively, without degradation in accuracy compared to the fully quantized INT8 baselines.  

%Specifically, on tested hardware it allows for improvement of up to $4.8x$ and $3.44x$ for latency and model size performance objectives without degradation in accuracy.

%Specifically for certain networks we demonstrate an improvement of up to 4.8x and 3.44x for latency and model size respectively, without degradation in accuracy compared to the fully quantizted INT8 baselines.  
\end{abstract}

%%%%%%%%%%%%%%%%%%%%%%%%%%%%%%%%%%%%%%%%%%%%%%%%%%%

\section{Introduction}

Recently, using Neural Architecture Search (NAS) to generate optimal Deep Neural Network (DNN) architectures for both computer vision (CV) and Natural Language Processing (NLP) applications has become a popular solution to increase inference efficiency of deployed models \cite{benmeziane2021comprehensive}\cite{chitty2022neural}. This trend reflects the increasing demand for highly specialized and efficient DNNs. NAS methodologies offer a systematic approach to exploring the vast search spaces of possible architectures, with the goal of identifying the optimal network structure for a specific task. In CV, NAS has been instrumental in developing architectures like EfficientNet \cite{tan2019efficientnet} and NASNet \cite{zoph2018learning}, which have set new high scores in benchmarks for image classification and object detection tasks. Similarly in NLP, NAS-based models like Evolved Transformer \cite{so2019evolved} have demonstrated improvement in language understanding and translation tasks. However, the growing complexity and cost of model's design and deployment necessitate an additional focus on model efficiency, especially in scenarios with computational and/or energy use constraints. This need has brought model efficiency techniques like quantization into focus.

One of the methods to achieve competitive accuracy in a quantized network is Post Training Quantization. This has been a deeply studied subject, but most of the research focuses on a quantization policy alone (\cite{yuan2022ptq4vit}) and a single optimization objective, which may lead to a sub-optimal solutions.

On the other hand, the joint neural architecture and quantization policy search introduces significant challenges. The task of jointly optimizing both the model's structure and numerical precision not only adds complexity but also increases computational demand, further expanding the already extensive search space. This has been to some extent addressed by Once For All (OFA)\cite{cai2019once}, showcasing the potential benefits of comprehensive optimization strategies. One of OFA's key contributions is the introduction of super-networks - large neural networks that can be dynamically adjusted to create smaller, specialized models without retraining. This is done with a novel weight-sharing approach that decouples the model training and architecture search stage to mitigate the computational overhead of sub-network architecture evaluation during NAS caused by training and validation cycles.

The one-shot super-network allows for flexible changes in depth, width, kernel size, and resolution without retraining, ideal for deployment on devices with varying computational capabilities. However, training these versatile OFA super-networks is complex and resource-intensive.

This idea has been extended to include quantization embedding in a joint Architecture, Pruning, and Quantization (APQ) \cite{wang2020apq} work. This facilitates the transfer of insights from the full-precision architecture predictor to the quantization-aware predictor and use it in conjunction with evolutionary search algorithm to identify best architecture and quantization policy. However, this approach is not cost effective due to complex training procedure of the FP32 super-network and need to generate a large dataset of architecture, quantization policy and network's accuracy data points to train the predictors. Additionally, this method was designed with convolutional neural networks(CNNs) in mind and is not yet proven to be transferable to Transformer-based models.

In the methods described above, there are some limitations, two of which we address in this work by proposing a systematic 2-step approach. The first problem is the search space that is tied to the super-network model. These models are hard to train and are both resource and time-consuming. To this end, this work leverages super-networks created with InstaTune \cite{Sridhar_2023_ICCV}. The second problem is search efficiency, which this work addresses by leveraging LINAS search algorithm \cite{cummings2022hardware}.

%%%%%%%%%%%%%%%%%%%%%%%%%%%%%%%%%%%%%%%%%%%%%%%%%%%

\section{Related Work}
 
 Currently, many joint quantization policy and neural architecture search methods have been proposed, which can be roughly categorized into two groups: quantization-aware training (QAT) and post-training quantization (PTQ).

 \textbf{QAT} methods generally incorporate quantization policy into the search space during the model training.
 For example, the bitwidth adaptive QAT \cite{youn2022bitwidth} proposes to redefine a meta-leaning task with bit widths to be applied in QAT.
 FLIQS \cite{dotzel2023fliqs} employs a reinforcement learning (RL) controller to propose per-layer formats and channel widths in quantization and NAS space during model training.

 Other efforts combine QAT with NAS using super-networks and differentiable NAS.
 For instance, DNAS \cite{wu2018mixed} proposes a differentiable super-network for QAT in which each layer contains several parallel quantized weights and activations with different precisions.
 EDMIPS \cite{cai2020rethinking} creates linear combination of branches for each bit-width, and then alternates training the layer weights and the branch weights.
 Q-BERT \cite{Zhao_2021} simultaneously conducts quantization and pruning at the subgroup-wise levels, and leverages differentiable NAS to assign scale and precision for parameters in each subgroup automatically.
 BatchQuant \cite{bai2021batchquant} employs batch statistics to adjust to changes in the activation distribution caused by the selection of quantized sub-networks. However, training QAT based super-networks can be extremely time consuming and computationally expensive.

 \textbf{PTQ} methods decouple the model training and quantization, which enables efficient transfer of the pre-trained model into a low-precision floating point or integer representation.
 ReLeQ \cite{elthakeb2018releq} performs PTQ quantization searches that utilize RL to allocate bit widths based on the model accuracy and cost estimates.
 HAWQ \cite{dong2019hawq} further explores PTQ searches to automatically select the relative quantization precision of each layer based on the Hessian spectrum.
 Recently, transformer-based models have been impressive in the computer vision field, which requires a lot of compute resources due to the self-attention module with quadratic time complexity.
 PTQ for transformer \cite{NEURIPS2021_ec895663} presents an efficient algorithm for quantization policy search to reduce the memory storage and computation costs of the vision transformer.
 PTQ4ViT \cite{yuan2022ptq4vit} proposes the twin uniform quantization method to reduce the quantization error on activation values by using a Hessian-guided metric to improve the accuracy of calibration. However, these PTQ based approaches for Transformers mainly focus on finding the best quantization policy and do not perform a joint search with the network architecture. 
 In contrast to these approaches, we propose a PTQ based technique using multi-objective evolutionary search algorithms incorporated with lightly trained predictors to search for sub-network architecture and corresponding quantization policy. Our method performs well across transformer-based architectures as well as CNN architectures.

%%%%%%%%%%%%%%%%%%%%%%%%%%%%%%%%%%%%%%%%%%%%%%%%%%%

\section{Methodology}
Most one-shot NAS approaches involve the creation of a trained super-network followed by a multi-objective evolutionary search to find optimal sub-networks.
Training super-networks often requires a manual specification of architectural elastic parameters and time-consuming and computationally intensive 
training methods like progressive shrinking \cite{cai2019once}, which becomes in-feasible especially for large models. Techniques like InstaTune\cite{Sridhar_2023_ICCV} leverage off-the-shelf pre-trained models and creates the elastic super-network during the downstream fine-tuning stage and improves the overall efficiency. Our approach is flexible and leverages super-networks generated by both InstaTune and other conventional methods like OFA. InstaTune offers two options based on compute resources to train the super-network. We employ the high cost option involving the use of a fully fine-tuned teacher network for knowledge distillation during fine-tuning.

\begin{table}[!h]
\caption{Search space size comparison for architecture ($S^{A}$), quantization policy ($S^{Q}$) and joint architecture and quantization policy ($S^{J}$) optimizations.}
\centering
{
\begin{tabular}{|c|c|c|c|c|}
\hline
Model        & $S^{A}$     & $S^{Q}$    & $S^{J}$      \\ \hline
BEiT-3       & $10^{13}$ & $10^{21}$ & $10^{34}$ \\ \hline
ViT          & $10^{13}$ & $10^{22}$ & $10^{35}$ \\ \hline
BERT         & $10^{13}$ & $10^{23}$ & $10^{36}$ \\ \hline
OFA ResNet50 & $10^{13}$ & $10^{37}$ & $10^{50}$ \\ \hline
\end{tabular}
}
\label{tab:search_space_size}
\end{table}

For multi-objective search on super-networks, we use the Lightweight Iterative NAS (LINAS) algorithm proposed in \cite{cummings2022hardware}. LINAS uses an iterative predictor-based approach to accelerate the search towards a near-optimal Pareto frontier, which is especially useful for vast search spaces defined by the joint architecture and quantization policy parameters (Table \ref{tab:search_space_size}).LINAS uses simple predictors such as ridge regression or support vector machine regression (SVR) and requires minimal data samples for accurate prediction as demonstrated in \cite{cummings2022hardware}. 

\textbf{Joint Architecture and Quantization Policy Search:}
Our proposed approach performs a joint architecture and quantization policy search to simultaneously optimize for network architecture and a corresponding quantization policy. We employ the LINAS search algorithm to explore a combined search space that includes both the network parameters and quantization bit widths for weights and activations. The combined search space size shown in Table \ref{tab:search_space_size} ranges from $10^{34}$ to $10^{50}$ for different models. We then perform post-training static quantization (PTQ static) for sub-networks with the selected mixed precision quantization policy - namely INT8 and FP32 - using the Intel Neural Compressor library. Static PTQ quantizes weights and activations of the model and requires a small representative calibration dataset to determine optimal quantization parameters for the activations.  While a variety of numerical precisions, such as BFloat16 \cite{kalamkar2019study} or INT4 could theoretically be considered for weights and activations to further extend set of possible solutions, the absence of readily accessible hardware platforms that natively support these format necessitated the limitation of our exploration to the combination of INT8 and FP32. This decision aligns our research with the capabilities of widely available hardware, ensuring the applicability and relevance of our findings.

\begin{table}[!h]
\caption{Parameters used to define architecture of the Transformer-based super-networks: number of layers (\textbf{$L^{A}$}), number of attention heads (\textbf{$H^{A}$}),intermediate sizes (\textbf{$Dffn^{A}$}). \textbf{$S^{A}$} is the approximated search space size for architecture-only optimization.}
\centering
{
\begin{tabular}{|c|c|c|c|c|}
\hline
Model  & $L^{A}$                        & $H^{A}$              & $Dffn^{A}$                 \\ \hline
BEiT-3 & 1x{[}11, 12{]}                 & 12x{[}6, 8, 10, 12{]} & 12x{[}2048, 2560, 3072{]} \\ \hline
ViT    & 1x{[}10, 11, 12{]}             & 12x{[}6, 8, 10, 12{]} & 12x{[}1024. 2048, 3072{]} \\ \hline
BERT   & 1x{[}6, 7, 8, 9, 10, 11, 12{]} & 12x{[}6, 8, 10, 12{]} & 12x{[}1024, 2048, 3072{]} \\ \hline
\end{tabular}
}
\label{tab:transformers_elastic_params}
\end{table}

\begin{table}[!h]
\caption{Parameters used to define architecture of the OFA ResNet50 super-network: depth ($D^{A}$), expand ratio ($E^{A}$) and width multiplier ($W^{A}$).}
\centering
{
\begin{tabular}{|c|c|c|c|}
\hline
Model        & $D^{A}$         & $E^{A}$                  & $W^{A}$                \\ \hline
OFA ResNet50 & 5x{[}0, 1, 2{]} & 18x{[}0.2, 0.25, 0.35{]} & 6x{[}0.65, 0.8, 1.0{]} \\ \hline
\end{tabular}
}
\label{tab:ofa_elastic_paramas}
\end{table}

In this work, we  explore unimodal and multimodal Transformer models like BEiT-3 \cite{wang2023image}, ViT \cite{dosovitskiy2020image}, and BERT \cite{devlin2018bert}. For Transformer-based architectures, we mainly use an elastic architecture search space with the number of layers, the number of attention heads, and the intermediate MLP dimensions (Table \ref{tab:transformers_elastic_params}). For completeness, we also include results on the pre-trained OFA \cite{cai2019once} ResNet50 \cite{he2015deep} super-network, with the architecture defined with a set of parameters presented in Table \ref{tab:ofa_elastic_paramas}.

\section{Experiments}

\begin{figure*}[h]
    \centering
    \begin{subfigure}{0.24\textwidth}
      \centering
      \includegraphics[width=\linewidth]{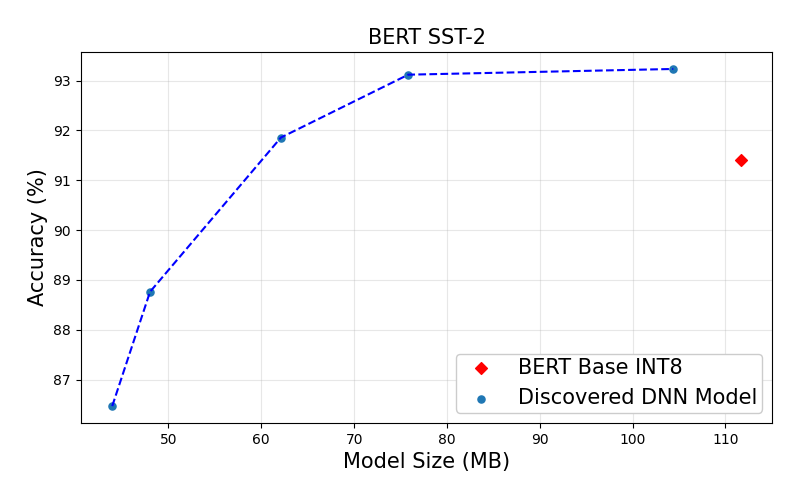}
      \caption{BERT model size}
      \label{fig:bert_model_size}

    \end{subfigure}
    \begin{subfigure}{0.24\textwidth}
      \centering
      \includegraphics[width=\linewidth]{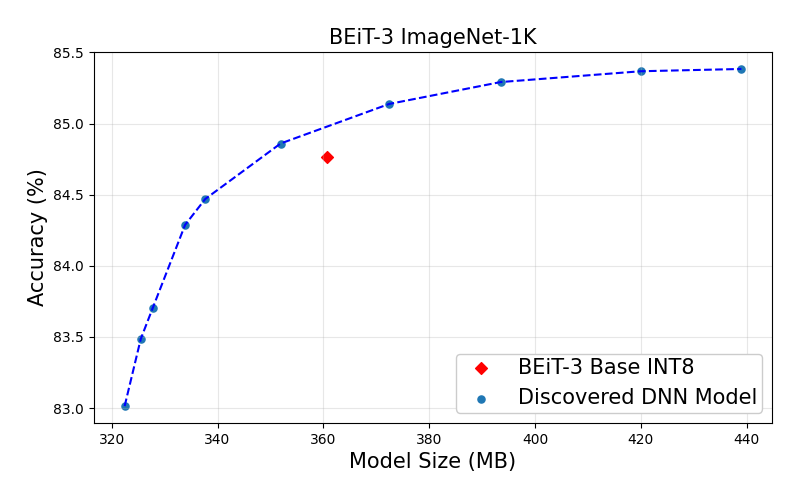}
      \caption{BEiT-3 model size}
      \label{fig:beit_model_size}
    \end{subfigure}
    \begin{subfigure}{0.24\textwidth}
      \centering
      \includegraphics[width=\linewidth]{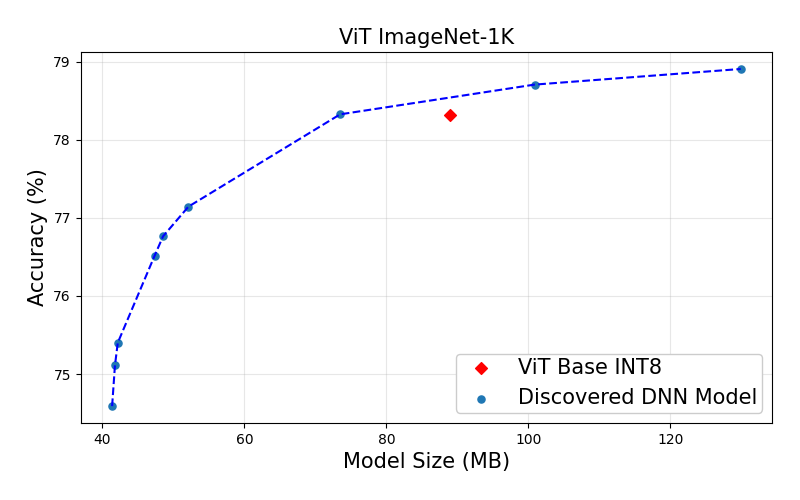}
      \caption{ViT model size}
      \label{fig:vit_model_size}
    \end{subfigure}
    \begin{subfigure}{0.24\textwidth}
      \centering
      \includegraphics[width=\linewidth]{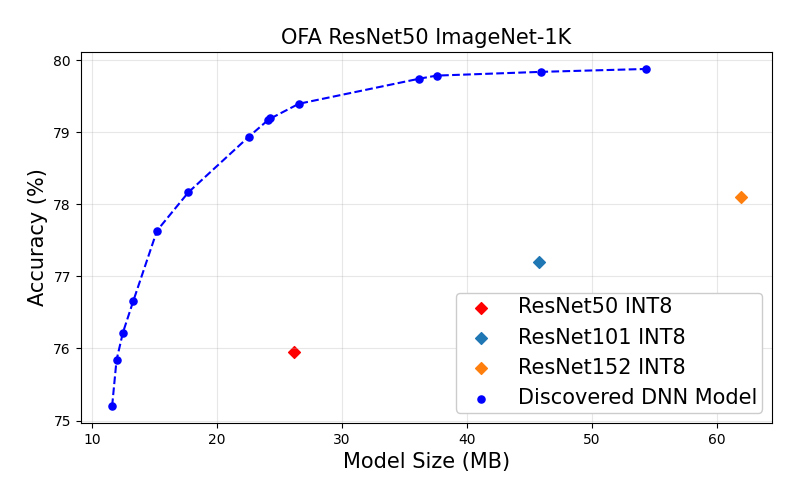}
      \caption{OFA ResNet50 model size}
      \label{fig:ofaresnet50_model_size}
    \end{subfigure}
    \bigskip
    \begin{subfigure}{0.24\textwidth}
      \centering
      \includegraphics[width=\linewidth]{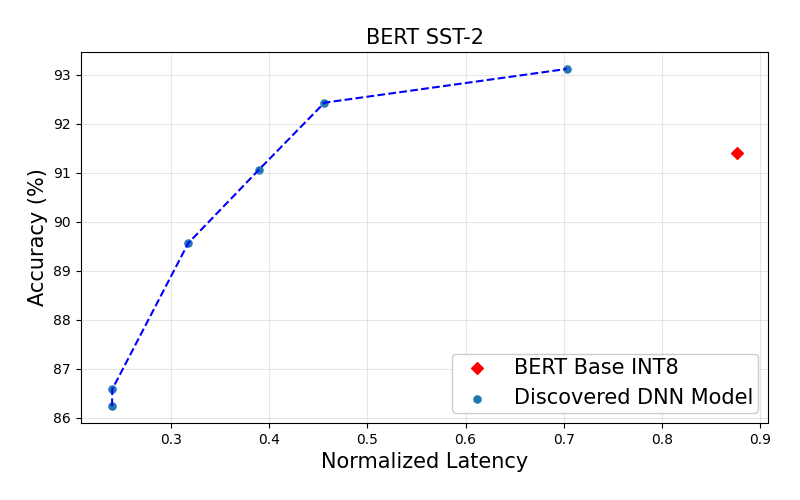}
      \caption{BERT latency}
      \label{fig:bert_latency}
    \end{subfigure}
    \begin{subfigure}{0.24\textwidth}
      \centering
      \includegraphics[width=\linewidth]{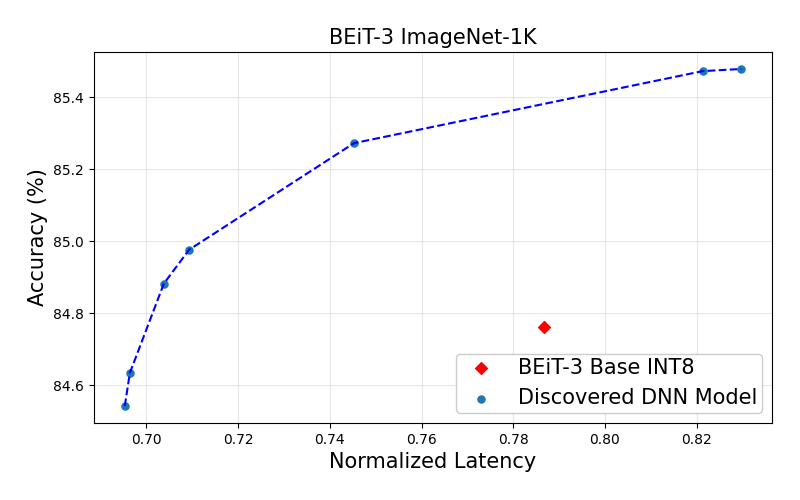}
      \caption{BEiT-3 latency}
      \label{fig:beit_latency}
    \end{subfigure}%
    \begin{subfigure}{0.24\textwidth}
      \centering
      \includegraphics[width=\linewidth]{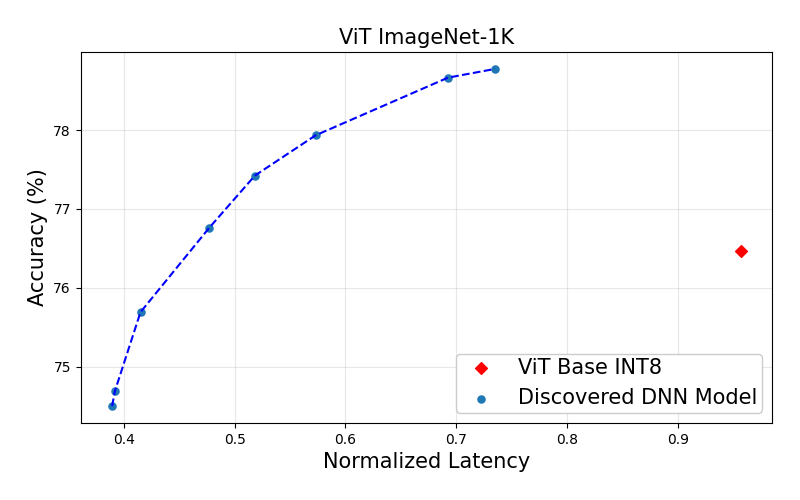}
      \caption{ViT latency}
      \label{fig:vit_latency}
    \end{subfigure}
    \begin{subfigure}{0.24\textwidth}
      \centering
      \includegraphics[width=\linewidth]{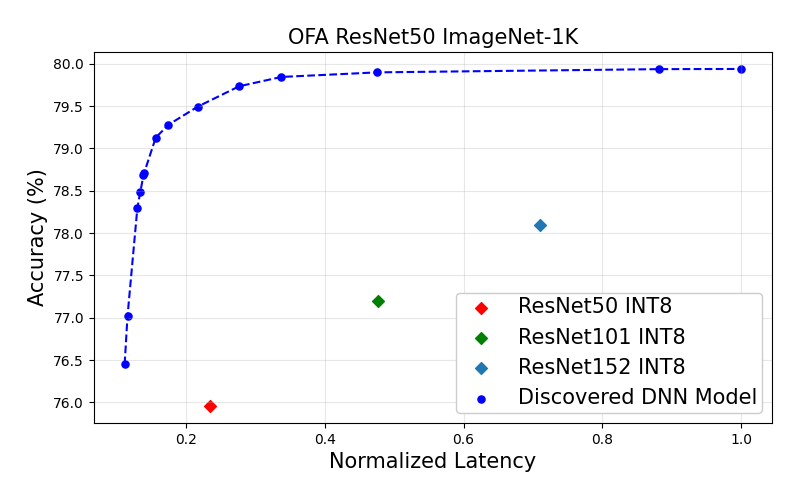}
      \caption{OFA ResNet50 latency}
      \label{fig:ofaresnet50_latency}
    \end{subfigure}
    \caption{ Joint architecture and quantization policy search Pareto front on BERT Base, BEiT-3 Base, ViT Base and OFA ResNet50 super-networks using model size and normalized latency as the search objectives.Each point shown on the Pareto front represents a discovered sub-network architecture and a corresponding quantization policy. } 
     \label{fig:pareto_fronts}
\end{figure*}

This section evaluates the proposed approach through experiments on diverse models suited for various tasks and domains. We tested ResNet50 and ViT for image classification, BERT for sentiment analysis in NLP, and BEiT-3, a multi-modal foundation model, also for image classification. These models were selected to demonstrate the applicability of our approach across different domains and architectural designs.
In our experiments, we showcased our method's flexibility through different super-network creation techniques. For BEiT-3 and BERT super-networks were built using InstaTune, involving elastic fine-tuning of pre-trained models on ImageNet-1K \cite{imagenet15russakovsky} for image classification and SST2 \cite{socher-etal-2013-recursive} for sentiment classification, respectively. Conversely, the ViT super-network was fully trained from scratch on ImageNet-1K to diversify super-net methodology, as it shares common elements with BEiT and BERT. For ResNet50, we used the OFA super-network trained on ImageNet-1K, demonstrating our approach's adaptability to various super-network training methods.
For each super-network a pair of multi-objective search experiments was performed. The primary metric was model accuracy on respective dataset, supplemented by either latency (single batch size) or model size (total memory occupancy). Latency measurement was performed on different Intel Xeon platforms (8280, 8360Y, 8380 and 8480) and was normalized.
Fig. \ref{fig:pareto_fronts} presents results in form of Pareto optimal fronts, showcasing performance enhancements from architectural and numerical optimizations. Each set of results includes reference point based on a baseline model with all linear layers in INT8.

The BERT model demonstrates a notable $2.20x$ speed increase at the same accuracy level compared to the INT8 baseline (Fig. \ref{fig:bert_latency}). Even the slowest identified configuration surpasses the baseline by $1.23x$ in speed, coupled with a $1.70\%$ absolute accuracy improvement. In terms of model size, our approach achieves a $1.83x$ reduction at equivalent accuracy levels. The highest accuracy network shows a $1.06x$ size improvement (Fig. \ref{fig:bert_model_size}).

For the BEiT-3 model, iterative improvements were noted. A model $1.02x$ smaller than the baseline was found, maintaining the same accuracy (Fig \ref{fig:beit_model_size}. The highest-scoring model showed $0.64\%$ accuracy improvement but was $1.22x$ larger than the baseline. Regarding latency, we found a model $1.12x$ faster than the baseline with comparable accuracy (Fig. \ref{fig:beit_latency}). Sub-model with the highest accuracy, improving by $0.69\%$, was $1.06x$ slower than the baseline.

For the ViT the biggest improvement can be seen with the latency objective (Fig. \ref{fig:vit_model_size}). Without sacrificing accuracy the search was able to identify a configuration that achieves 1.97x improvement over baseline. Sub-network with the highest top1 accuracy $78.77\%$ ($2.30\%$ absolute improvement over the baseline) also offers improvements in latency ($1.20x$). For the model size objective (Fig. \ref{fig:vit_latency}) the improvement between reference baseline and on-par accuracy-wise sub-network is $42.40\%$, and when considering fixed model size, the identified sub-model yields improved accuracy by $2.03\%$ over the baseline.

Joint search on OFA ResNet50 offers significant improvements for both performance objectives. Specifically, without accuracy loss model size was reduced by $2x$, $3.06x$ and $3.44x$ for ResNet50, ResNet101 and ResNet152 baselines (Fig. \ref{fig:ofaresnet50_model_size}). Similarly, for the latency objective, an improvement of $2.30x$, $4.36x$ and $4.8x$ was observed (Fig. \ref{fig:ofaresnet50_latency}).

\subsection{Ablations on Quantization only search}
\begin{figure*}[h]
    \centering
    \begin{subfigure}{0.35\textwidth}
      \centering
      \includegraphics[width=\linewidth]{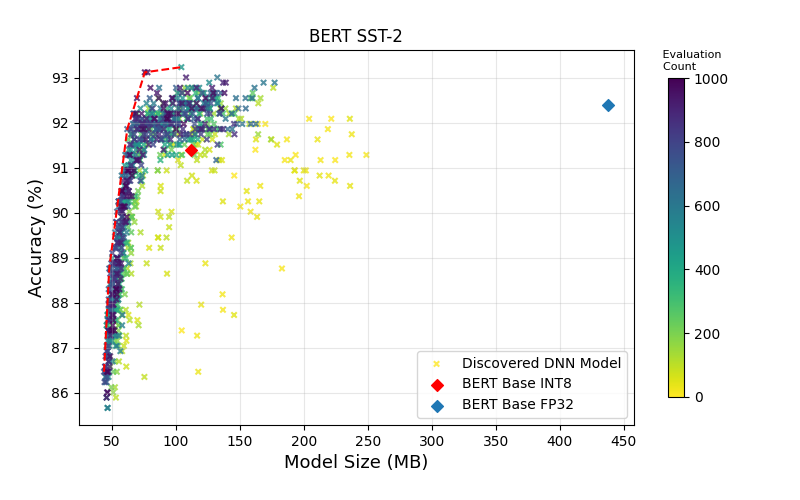}
      \caption{BERT SST-2 model size}
      \label{fig:qp_search_preogression_bert}
    \end{subfigure}
    \begin{subfigure}{0.35\textwidth}
      \centering
      \includegraphics[width=\linewidth]{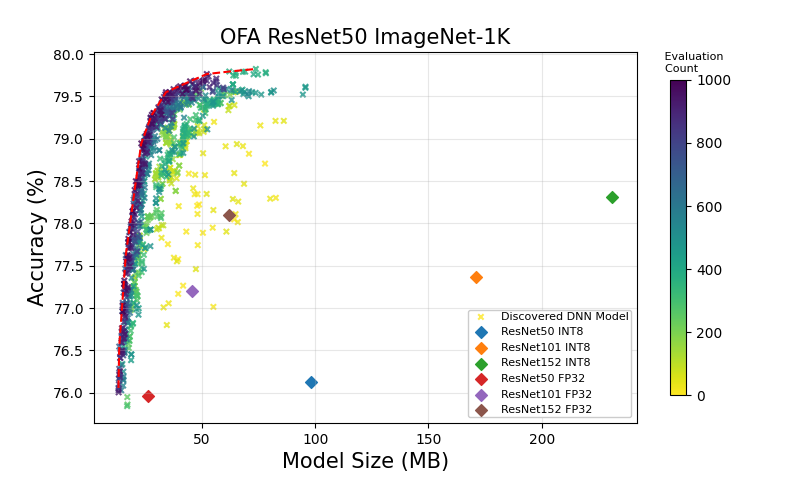}
      \caption{OFA ResNet50 model size}
      \label{fig:qp_search_preogression_ofa}
    \end{subfigure}
    \begin{subfigure}{0.273\textwidth}
      \centering
      \includegraphics[width=\linewidth]{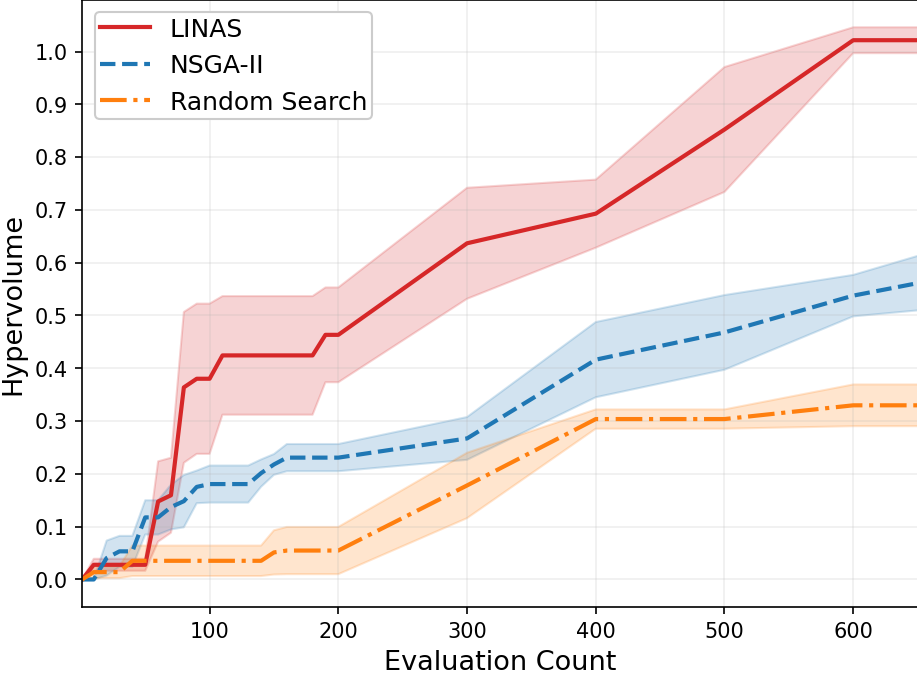}
      \caption{BERT SST-2 hypervolume}
      \label{fig:qp_search_preogression_hv}
    \end{subfigure}
    \caption{Search progression using LINAS for BERT Base (\ref{fig:qp_search_preogression_bert}) and OFA ResNet50  (\ref{fig:qp_search_preogression_ofa}) super-networks. A clear progression towards the near-optimal Pareto front can be seen with an increase in evaluation count. Additionally, a significant improvement in model size is observed when compared to the FP32 baselines. Figure\ref{fig:qp_search_preogression_hv} represents the BERT Base hypervolume progression for different search algorithms is presented in \ref{fig:qp_search_preogression_hv}}
     \label{fig:qp_search_preogression}
\end{figure*}

\begin{figure}[h]
    \centering
    \includegraphics[width=0.5\columnwidth]{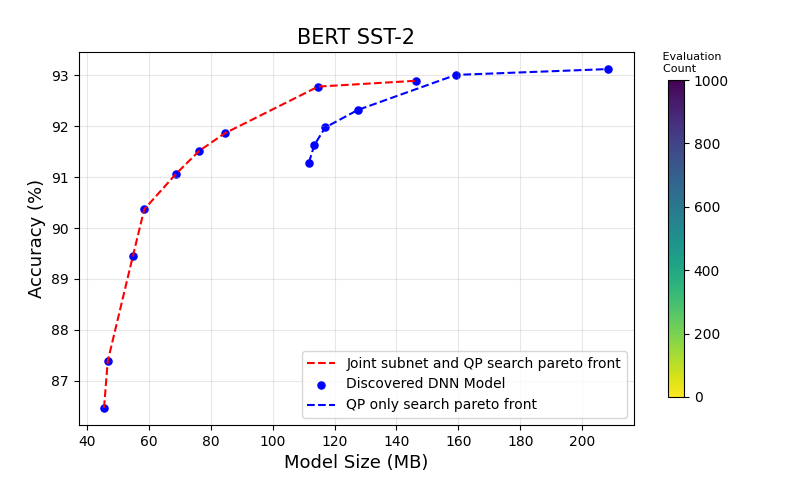}
    \caption{Comparison of quantization policy only search and joint architecture and quantization policy search with accuracy and model size on BERT Base fine-tuned with InstaTune on SST2 dataset.}
    \label{fig:search_comparison}
\end{figure}

Fig. \ref{fig:search_comparison} shows the comparison between quantization policy only search and joint sub-network and quantization policy search on the BERT SST2 super-network. The results indicate that performing joint search using our approach significantly improves the model size for similar accuracy, when compared to quantization policy only search.
Figures \ref{fig:qp_search_preogression_bert} and \ref{fig:qp_search_preogression_ofa} demonstrates a drastic improvement in the search progression using LINAS with increasing evaluation counts.
Additionally, multiple search experiments were performed with various search algorithms - LINAS, NSGA-2 and Random Search - to validate efficiency of the selected search algorithm. Averaged results with standard deviation as a plane prove efficacy of LINAS on the joint search space and are presented as a hypervolume in Fig. \ref{fig:qp_search_preogression_hv} for BERT SST2. 

%%%%%%%%%%%%%%%%%%%%%%%%%%%%%%%%%%%%%%%%%%%%%%%%%%%

\section{Conclusions}
In this work we demonstrate the effectiveness of a joint architecture and quantization policy search in optimizing DNN models for varied tasks. Our approach, which simultaneously optimizes the network architecture and quantization policy has shown considerable improvements in efficiency without compromising on the accuracy. Specifically, for models like BERT, BEiT3, ViT and ResNet50, we observed a substantial enhancement in terms of latency, model size and classification accuracy. By focusing on widely supported numerical precisions, like INT8 and FP32, we ensured that our findings are applicable and relevant to a broad range of real-world scenarios.
Future research in this area could explore integrating a broader range of numerical precisions, such as BFloat16 and INT4, as hardware becomes more readily available to researchers. Another avenue for future work involves researching applicability of the optimization methodology described in this work across a broader set of domains, extending beyond image and sentiment classification. Investigating these optimizations to current frontiers of research like large language models is also an important research vector for this work.

\bibliographystyle{unsrt}  
\bibliography{main}

\end{document}